\documentclass[10pt,twocolumn,letterpaper]{article}

\usepackage{cvpr}              % To produce the CAMERA-READY version
\usepackage{multirow}
\usepackage{pifont}
\usepackage{algorithm}
\usepackage{algpseudocode}
\usepackage{graphicx}
\usepackage{makecell}
\usepackage{soul}
\usepackage{comment}
\usepackage{xcolor}
\usepackage{bbm}
\usepackage[accsupp]{axessibility}  % Improves PDF readability for those with disabilities.
\usepackage[dvipsnames]{xcolor}
\usepackage{xspace}

\definecolor{cvprblue}{rgb}{0.21,0.49,0.74}
\usepackage[pagebackref,breaklinks,colorlinks,citecolor=cvprblue]{hyperref}

\def\paperID{353} % *** Enter the Paper ID here
\def\confName{CVPR}
\def\confYear{2025}

\author{Ziyi Liu\\
University of Science and Technology Beijing\\
{\tt\small liuziyi.iair@gmail.com}
\and
Yangcen Liu\\
Georgia Institute of Technology\\
{\tt\small yliu3735@gatech.edu}
}

\title{Bridge the Gap: From Weak to Full Supervision for Temporal Action Localization with PseudoFormer} % 

\begin{document}

\maketitle

\begin{abstract}

Weakly-supervised Temporal Action Localization (WTAL) has achieved notable success but still suffers from a lack of temporal annotations, leading to a performance and framework gap compared with fully-supervised methods. 
While recent approaches employ pseudo labels for training, three key challenges: generating high-quality pseudo labels, making full use of different priors, and optimizing training methods with noisy labels remain unresolved.
Due to these perspectives, we propose \textbf{PseudoFormer}, a novel two-branch framework that bridges the gap between weakly and fully-supervised Temporal Action Localization (TAL).
We first introduce RickerFusion, which maps all predicted action proposals to a global shared space to generate pseudo labels with better quality.
Subsequently, we leverage both snippet-level and proposal-level labels with different priors from the weak branch to train the regression-based model in the full branch.
Finally, the uncertainty mask and iterative refinement mechanism are applied for training with noisy pseudo labels.
PseudoFormer achieves state-of-the-art WTAL results on the two commonly used benchmarks, THUMOS14 and ActivityNet1.3.
Besides, extensive ablation studies demonstrate the contribution of each component of our method.
\end{abstract}
\section{Introduction}
% 目前的弱监督已经取得了长足的进步，但是实际上WTAL整体上使用的MIL的思路，和TAL的方法以及性能差距非常大。虽然有PivoTAL提出了regression model学习伪标签的方法，但是如何更加合理的bridge TAL和WTAL有待研究。

Weakly-supervised Temporal Action Localization (WTAL), which aims to use video-level annotations to identify and localize temporal action instance, has gained significant traction in recent years~\cite{autoloc,cascade,actionunit,weaklysuperviseduncertainty,adaptivetwostream}. 
Despite the progress, a substantial gap remains between weakly-supervised and fully-supervised TAL methods, in both the framework and performance.
WTAL methods predominantly rely on Multiple Instance Learning (MIL)~\cite{mil} for localization-by-classification, while TAL approaches often employ regression models to learn from densely annotated proposals.
Although recent work~\cite{pivotal} has explored using regression models to learn from generated labels, a more comprehensive approach to effectively bridge the gap of framework and performance between WTAL and TAL remains an open challenge.
\begin{figure}
  \centering
       \centering
        \includegraphics[width=1.0\linewidth]{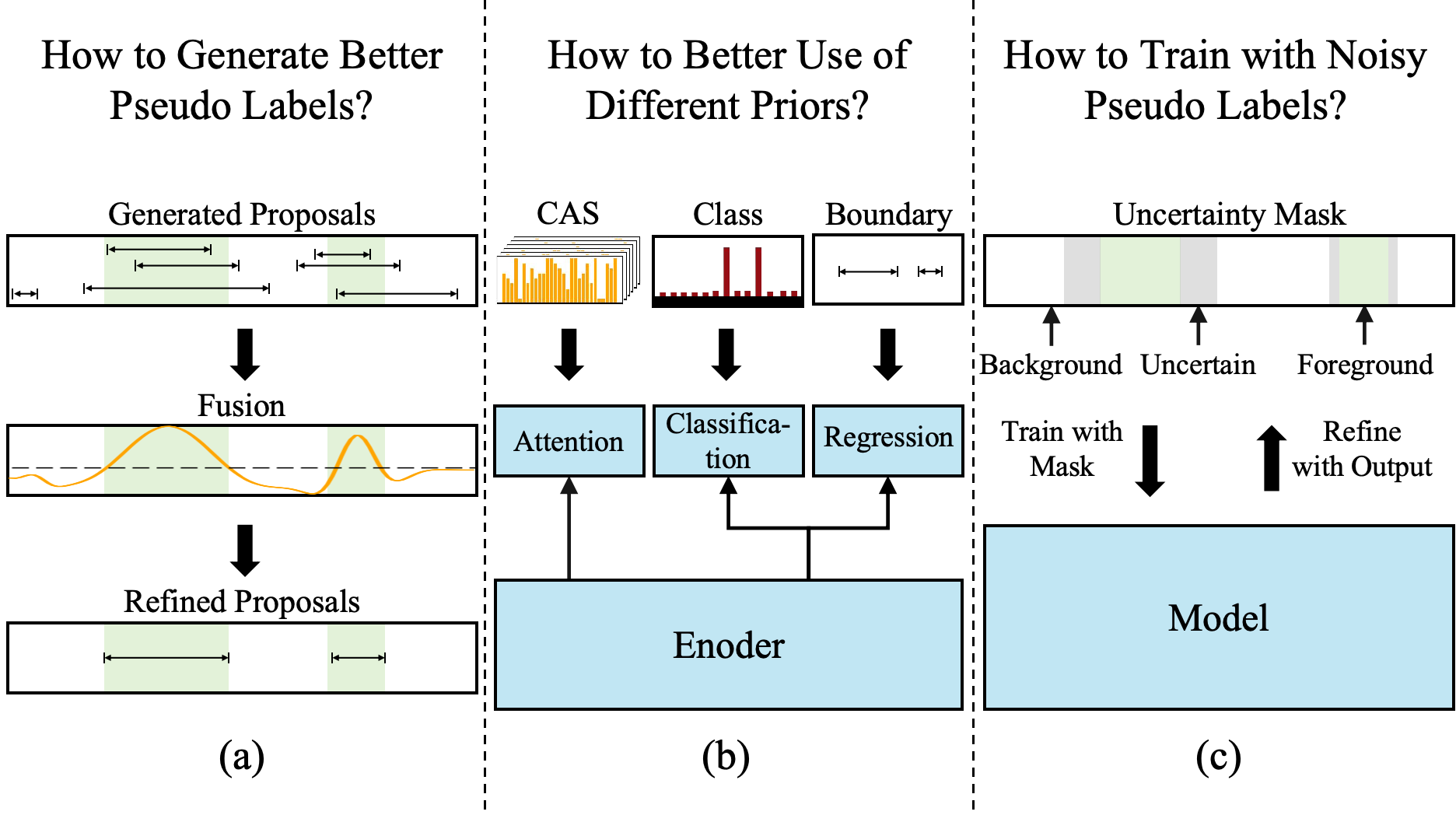}
       \caption{Our paper aims at addressing the three primary questions: (a) How to improve the quality of generated labels from the base model? (b) What priors could the regression model learn from, and how could it better use them? (c) How to train with noisy labels in uncertainty?}
       \label{topright}
\end{figure}

Some of the latest approaches~\cite{gaussfusion,pivotal} design two-stage schemes to generate pseudo labels to train a regression-based model.
The introduction of the pseudo-label learning framework boosts performance significantly.
Unlike previous WTAL methods~\cite{co2net,delu,asmloc,acsnet}, the key to these methods is to improve the quality of pseudo labels with post-processing strategies.
However, current approaches face several limitations: they often fail to fully leverage the priors generated in the first stage, leading to suboptimal use of valuable contextual information.
Furthermore, they lack effective integration with existing fully-supervised TAL frameworks, which can hinder the adaptation and consistency of the methods.
We propose a two-branch (weak and full) \textbf{PseudoFormer}, focusing on addressing the following three critical problems shown in~\cref{topright}:
(1) \emph{How can we generate pseudo labels with confidence scores and more reliable boundaries?}
(2) \emph{How can we fully leverage the priors contained in the generated pseudo labels?}
(3) \emph{How can we effectively train a regression model with noisy pseudo labels?}

To improve the quality of pseudo proposals, we introduce RickerFusion, a pseudo label generation pipeline designed to fuse outputs from a weak branch. The weak branch is functioned by a model trained using Multiple Instance Learning (MIL)~\cite{co2net,delu,ddgnet}.
Drawing inspiration from the Bird's Eye View (BEV)~\cite{bevf,bevdet} framework, we propose to fuse the temporal predictions from the weak branch in a shared space.
Specifically, RickerFusion conceptualizes the proposal space as a top-down, unified representation where confidence scores and boundary information are treated as spatial coordinates in this space.
We first follow AutoLoc~\cite{autoloc} applying the Outer-Inner-Contrastive (OIC) score from the snippt-level predictions (SPs) to compute the proposal confidence scores.
Subsequently, we map each output proposal to a Ricker wavelet~\cite{ricker} distribution and fuse these proposals into a unified global space. This fusion effectively consolidates predictions from diverse perspectives, allowing for noise reduction and enhanced mutual reinforcement among proposals, and ultimately capturing both the strength and spatial coherence of the detections.
The fused representation is then leveraged to generate the final fused proposals.
Similar to how BEV provides a global, consolidated view from multiple sources, RickerFusion can also integrate multi-scale perception into a cohesive output, improving both the accuracy and reliability of the pseudo proposals.

To effectively learn from diverse priors within pseudo labels, and bridge the gap between 
the frameworks between WTAL and FTAL methods, we propose a unified framework consists with a weak branch and a full branch, that learns from both video-level and proposal-level information.
In the full branch, we incorporate a regression-based model, similar to mainstream TAL architectures~\cite{actionformer, tridet}, to leverage the proposal-level pseudo labels generated in the weak branch. 
This allows the model to learn action boundaries and durations in a manner consistent with fully-supervised TAL methods.
Additionally, by learning from the SPs in the weak branch, the model enriches its encoded snippt-level features with video classification information.
Each pseudo label carries natural priors from the MIL process used in weak supervision, along with artificial priors introduced through design, enhancing the performance of the proposed method.

In the context of WTAL, the generated pseudo proposals are inherently noisy and lack the temporal accuracy compared to the ground-truth.
To alleviate the distraction caused by wrong labels when training with noisy pseudo proposals, we introduce an uncertainty mask to enhance the robustness of our method.
Recognizing that the boundaries of these proposals are likely to be less reliable compared to the centers, the mask allows us to exclude specific segments at the internal and external boundaries of each proposal from the training process. 
By initially focusing on the more reliable central regions, we mitigate the risk of incorporating noisy information that may distract the model. 
As training progresses and the model's confidence in the reliability of the proposals increases, we gradually reduce the excluded regions, eventually integrating all parts of the proposals. 
This adaptive approach ensures more effective training with noisy (pseudo) labels, progressively enhancing the model's ability to leverage the entire proposal information while minimizing the influence of uncertain boundary regions.

To summarize, our contribution is threefold:

\begin{itemize}
    \item We propose PseudoFormer, a novel two-branch method for the WTAL task, which bridges the gap in both framework and performance between Weakly and Fully-supervised Temporal Action Localization.
    \item Considering the three primary problems, PseudoFormer utilizes the RickerFusion strategy to generate high-quality pseudo labels, leverages both video and proposal level labels with different priors, and proposes an uncertainty mask with a refinement mechanism for learning with noisy labels.
    \item We conduct comprehensive experiments on two benchmarks THUMOS14~\cite{thumos} and ActivityNet1.3~\cite{activitynet}. 
    Our method achieves significant performance over existing WTAL methods.
\end{itemize}
\section{Related Work}

\begin{figure*}[t]
  \centering
       \centering
       \includegraphics[width=1.0\linewidth]{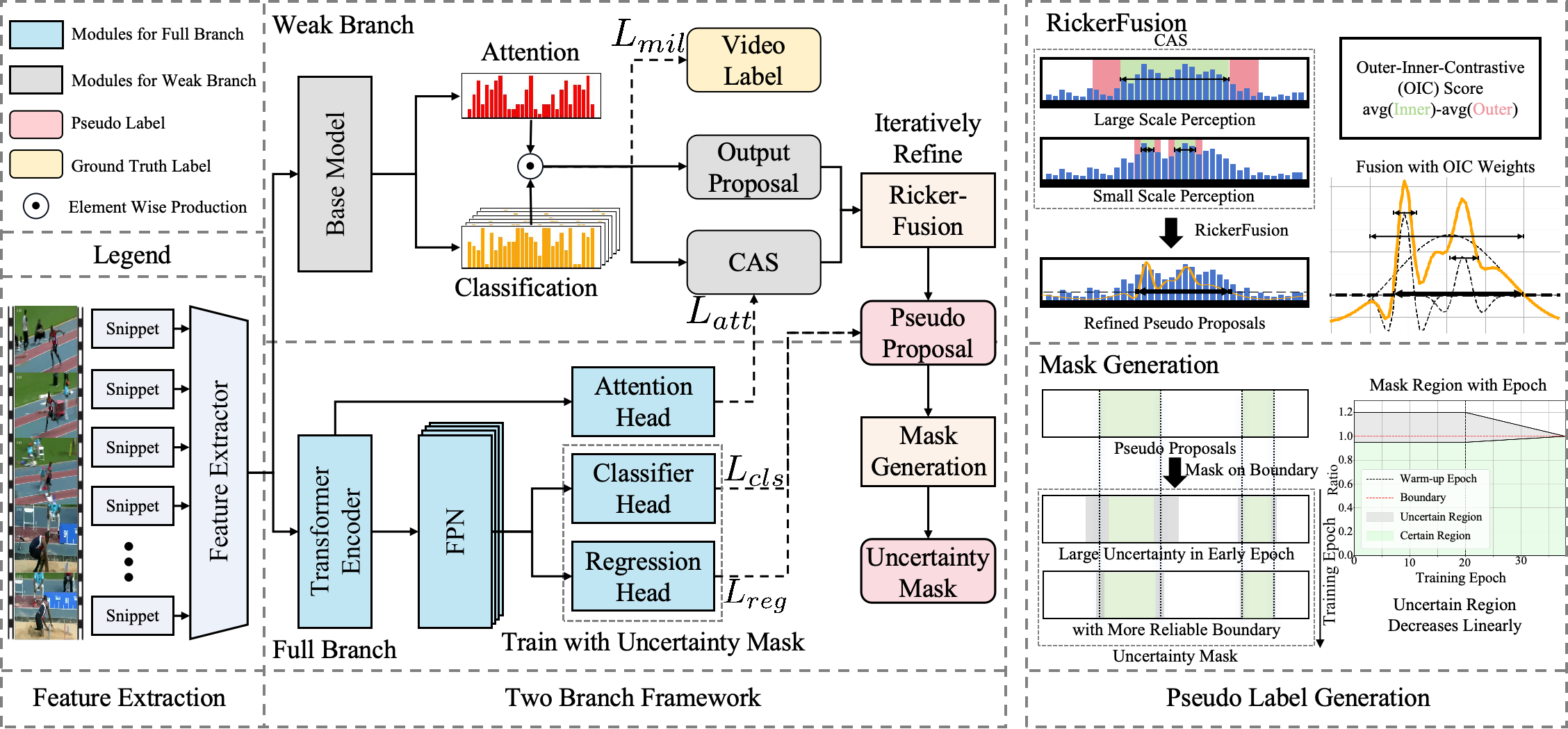}
       \caption{Overall framework of PseudoFormer. 
       (a) \textbf{Weak Branch}: After feature extraction, the base model predicts agnostic attention and classification scores. Using Multi-instance Learning (MIL) with video-level labels, it outputs proposals and the snippt-level predictions (SPs). 
       (b) \textbf{Full Branch}: A regression-based model is trained on the \emph{snippt-level predictions (SPs)}, \emph{pseudo proposals}, and \emph{uncertainty mask}, and is used for final inference. 
       (c) \textbf{RickerFusion}: To improve the quality of pseudo labels, RickerFusion maps predictions across perception scales into a shared space, producing better pseudo labels by fusing predictions from the weak branch. 
       (d) \textbf{Mask Generation}: To train with noisy pseudo labels, an uncertainty mask is applied to proposal boundaries, and the uncertain regions gradually decrease after the warm-up epoch. The pseudo proposals and the uncertainty mask are iteratively refined during training.}
       \label{fig:pipeline}
\end{figure*}

\noindent\textbf{Temporal Action Localization.}
Fully-supervised Temporal Action Localization (TAL) aims to simultaneously classify and localize action instances in untrimmed videos.
Fully-supervised approaches can generally be categorized into two groups: two-stage methods~\cite{dcan,learningsalient,temporalcontext,basictad} and one-stage methods~\cite{actionunit,seedsequence,stepbystep,learningdisentangled}.
Two-stage approaches consist of two steps: proposal generation and classification.
In contrast, one-stage methods unify proposal generation and classification into a single step. Following object detection methods~\cite{detr}, DETR-like decoder methods are proposed~\cite{end2endtal,react,relaxed,tallformer}. 
Recent advances, another series of work following ActionFormer~\cite{actionformer,tridet,dyfadet} have led to significant performance gains.
These methods build a multi-scale feature pyramid network followed by a classification and regression head.
In this work, we mainly follow the mainstream actionformer-based framework.

\noindent\textbf{Weakly-supervised Temporal Action Localization.}
To address the high cost for the snippet-level annotation, Weakly-supervised Temporal Action Localization (WTAL) has been widely studied~\cite{autoloc,,co2net,delu,ddgnet,gtal,contrastwtal,gao2025learning}.
In contrast to fully-supervised methods, UntrimmedNet~\cite{untrimmednets} was the first to adopt multi-instance learning~\cite{mil,mil-img} for action recognition, classifying salient segments to generate action proposals without a regression head.
A significant body of work~\cite{twostream,adaptivetwostream,co2net,delu,ddgnet} exploits the relationship between cross-modal channels. 
CO2-Net~\cite{co2net} applies a cross-modal attention mechanism to enhance feature consistency within the structure. 
DDG-Net~\cite{ddgnet} directly designs a residual graph structure module to construct the cross-modal connection. 
A more efficient and popular method is to guide the training with uncertainty learning strategy~\cite{weaklysuperviseduncertainty,ugct,delu,cascade}. 
DELU~\cite{delu} introduces evidential learning to guide the sampling strategy and learning.
Recently, a line of work~\cite{weaklyknoledge,pmil,asmloc,refineloc,twostream,pivotal} attempts to generate pseudo labels to iteratively guide the training.

\noindent\textbf{Learning from Pseudo Labels.}
Learning from snippet-level pseudo labels could serve as a bridge between TAL and WTAL.
When the example is not annotated, producing a pseudo label through confidence score thresholding is a commonly used approach in traditional methods~\cite{pseudodn,fixmatch,flexmatch, zhang2025moma}. 
Between the WTAL and TAL settings, for point-supervised (PTAL)~\cite{pgcn,hrpro,tspnet} and semi-supervised (SSTAL)~\cite{npl,boostingsemi,towardssemi} settings, several works focus on learning from the generated pseudo labels.
For PTAL, based on the point annotation distribution, TSP-Net~\cite{tspnet} proposes a center score learning method to dynamically predict the alignment from the saliency information.
For SSTAL, NPL~\cite{npl} iteratively generates pseudo labels with a ranking and filtering mechanism to improve the results.
For WTAL, P-MIL~\cite{pmil} uses a two-stage framework, using pseudo labels to guide the training.
Zhou et al.~\cite{gaussfusion} design a Gaussian-based self-correction method to refine the boundary and score biases, thus producing pseudo labels with higher quality on the action level. 
PivoTAL~\cite{pivotal} incorporates prior knowledge to generate informative proposals for regression models to learn.
However, how to generate pseudo labels with better quality, how to make full use of different priors, and how to train with the noisy pseudo labels with uncertainty, still remain to be explored.
\section{Method}
In the Weakly-supervised Temporal Action Localization (WTAL) task, only video-level labels are provided, noted as $\mathbf{V}= \{ \boldsymbol{v}_i, \boldsymbol{y}_i \}_{i=1}^S$, where $S$ denotes the total number of video samples, $\boldsymbol{v}_i$ represents the $i$-th video, and $\boldsymbol{y}_i$ indicates the corresponding video-level category label among $C$ categories.
As shown in~\cref{fig:pipeline}, PseudoFormer is a two-branch framework composed of a Multiple Instance Learning (MIL)-based weak branch and a regression-based full branch. The weak branch employs MIL to train a base model, generating several useful priors: the snippt-level predictions (SPs) $\boldsymbol{Z}$ and proposal set $P$.
Through the RickerFusion and Mask Generation Modules, refined pseudo proposals $\hat{P}$ and an uncertainty mask $\boldsymbol{U}$ are produced, enhancing proposal quality. In the full branch, a regression-based model is trained using SPs $\boldsymbol{Z}$, the uncertainty mask $\boldsymbol{U}$, and pseudo proposals $\hat{P}$.
After training, only the full-branch part is used for inference.

\subsection{Weak Branch}

\noindent\textbf{Base Model.}
Following previous methods~\cite{untrimmednets, pmil}[ref], we use off-the-shelf RGB and optical flow snippet-level features as input, denoted as $\boldsymbol{F} \in \mathbb{R}^{T \times D}$, where $T$ denotes the number of snippets.
Given the snippet-level features, a category-agnostic attention branch $\mathcal{M}^a$ is used to compute the snippt-level attention predictions (SAPs) $\boldsymbol{\varphi} \in \mathbb{R} ^{T\times 1}$, and a classification branch is used to predict the snippt-level category predictions (SCPs)  $\boldsymbol{\Psi} \in \mathbb{R}^{T\times (C + 1)}$, where $C + 1$ denotes the number of categories plus one background class.

After that, based on the assumption that background is present in all videos but filtered out by the SAPs $\boldsymbol{\varphi}$, the original predicted video-level classification scores $\boldsymbol{\hat{y}}_{base} \in \mathbb{R}^{C+1}$, and the foreground video-level classification scores $\boldsymbol{\hat{y}}_{supp} \in \mathbb{R}^{C+1}$ suppressed by attention, are derived by applying a temporal top-k aggregation strategy $f_{agg}$~\cite{mil,untrimmednets}, respectively:

\begin{equation}
    \boldsymbol{\hat{y}}_{base} = f_{agg}(\boldsymbol{\Psi}),  \;
    \boldsymbol{\hat{y}}_{supp} = f_{agg}(\boldsymbol{\varphi} \odot \boldsymbol{\Psi}),
\label{equ:base_sup}
\end{equation}
where $\odot(\cdot)$ is the element-wise product.

To consider background information of each video, we extend the video-level label: $\boldsymbol{y}_{base} = [\boldsymbol{y}, 1] \in \mathbb{R}^{C+1}$ and $\boldsymbol{y}_{supp} = [\boldsymbol{y}, 0] \in \mathbb{R}^{C+1}$. Guided by the video-level category label $\boldsymbol{y} \in \mathbb{R}^{C}$, the multi-instance learning classification loss is defined as:

\begin{equation}
L_{mil} = -\sum_{c=1}^{C+1}(\boldsymbol{y}_{base}\log \hat{\boldsymbol{y}}_{base} + \boldsymbol{y}_{supp}\log \boldsymbol{\hat{y}}_{supp}),
\label{equ:label}
\end{equation}

\noindent\textbf{Generating Different Priors.} After training, the predicted snippt-level predictions (SPs), $\boldsymbol{Z} \in \mathbb{R}^{T \times (C+1)}$ is computed as $\boldsymbol{\varphi} \odot \boldsymbol{\Psi}$, which is iteratively thresholded to obtain the final localization results $P = \{(s_i, e_i, o_i, c_i)\}_{i=1}^K$ \cite{delu}.
$K$ is the total number of output proposals, and $s_i$, $e_i$, $o_i$, $c_i$ denote the start, end, confidence score, and class for each proposal $P_i$, respectively.
Following AutoLoc~\cite{autoloc}, we compute the score for each proposal with OIC score:
\begin{equation}
    o_i=OIC(P_i)=avg(P_i^{inner})-avg(P_i^{outer}),
\label{equ:oic}
\end{equation}
where $P_i^{inner}$ and $P_i^{outer}$ are the SPs values within and outside the proposal with a fixed ratio.
The OIC score effectively captures the boundary characteristics of the predicted proposals and provides a reliable measure of confidence.

After soft-NMS~\cite{softnms}, we obtain the output proposals for proposal-level annotation, and SPs $\boldsymbol{Z}$ for snippet-level annotation with different priors from the weak branch for the full branch training.

\subsection{Pseudo Label Generation}

\textbf{RickerFusion.} The output proposals $P$ from the base model are generated individually considering only snippet scores from SPs,
% vary in scale, with higher thresholds yielding smaller-scale perceptions.
% These proposals are generated individually considering only snippet scores from CAS in different perceptions, 
ignoring the overall relationship between different proposals and scales.
Additionally, the proposals are heavily overlapped due to the multi-level thresholding, which fail to accurately describe boundaries for the full branch training.
Inspired by Bird's Eye View (BEV)~\cite{bevdet}, RickerFusion addresses this by mapping all proposals to a unified space to get a global perception, enhancing spatial understanding and simplifying the representation. 
Specifically, we use the Ricker wavelet~\cite{ricker}, which emphasizes the center of each proposal while suppressing peripheral areas, enabling effective fusion of proposals across different scales. 
This approach enhances temporal action localization by achieving balanced and robust multi-scale proposal alignment, as shown in~\cref{fig:pipeline}.

For each proposal, we map it using the Ricker distribution as follows:
\begin{equation}
    \phi_i (t)=\frac{2}{\sqrt{3\sigma_i}\pi^{\frac{1}{4}}}(1-(\frac{t-m_i}{\sigma_i})^2)e^{-\frac{(t-m_i)^2}{2\sigma_i^2}},
\label{equ:ricker}
\end{equation}
where $\sigma_i=\frac{e_i-s_i}{2}$ denotes the length, and $m_i=\frac{s_i+e_i}{2}$ denotes the midpoint of $P_i$.
The Ricker wavelet naturally emphasizes the central part of the proposal while applying a suppression at the boundaries. For example, $\phi_i (t)$ achieves the maximum positive value when $t=m_i$, and $\phi_i (s_i)=\phi_i (e_i)=0$. Besides, $\phi_i(t)$ will be negative, if $t<s_i$ or $t>e_i$, indicating the locations out of the proposal are suppressed. This suppression effect initially increases away from the center before gradually tapering off, helping to reduce noise and minimize the influence of less reliable boundary regions. 
The characteristic aligns well with the localization results, as it ensures that each proposal captures core activity with high confidence, while the regions out of the boundaries should NOT contain the target action.

Then we fuse all proposals to a shared global space as follows:
\begin{equation}
    \Phi^l = \sum_{i=1}^{K} \phi_i o_i \cdot \mathbbm{1}(c_i = l),
\label{equ:ricker2}
\end{equation}
where $\Phi \in \mathbb{R}^{T\times C}$ denotes the final fused wavelet. 
Each proposal is weighted by its confidence score $o_i$, and predictions with incorrect classification categories are filtered out.

We obtain the final pseudo label $\hat{P} = \{ (\hat{s}_i, \hat{d}_i\, \hat{c}_i \}_{i=1}^{\hat{K}}$, from $\Phi$ with a threshold of $y=0$ for the activated area.
The pseudo proposal would have better quality and more reliable boundaries.

\noindent\textbf{Mask Generation.} Although the generated pseudo labels achieve reasonably high quality, a significant gap remains between these pseudo labels and the true ground truth labels. 
To improve training with these noisy pseudo labels, we leverage a prior assumption about proposal accuracy: the central regions of proposals tend to be more reliable, while the boundaries are more uncertain. 
Based on this observation, we introduce the uncertainty mask as follows:
\begin{align}
& \mathcal{M}_i(t) = 
\begin{cases}
0, & \text{if } t \in \mathcal{I}_1 \cup \mathcal{I}_2, \\
1, & \text{otherwise}.
\end{cases} \notag \\
\mathcal{I} = & \left\{
\begin{aligned}
& \mathcal{I}_1 = \left[ \hat{s}_i - \alpha (\hat{e}_i - \hat{s}_i), \hat{s}_i + \beta (\hat{e}_i - \hat{s}_i) \right], \\
& \mathcal{I}_2 = \left[ \hat{e}_i - \beta (\hat{e}_i - \hat{s}_i), \hat{e}_i + \alpha (\hat{e}_i - \hat{s}_i) \right].
\end{aligned}
\right.
\end{align}
where $\mathcal{M} \in \mathbb{R}^{T}$ denotes the uncertainty mask for $\hat{P}_i$, while $\alpha$ and $\beta$ denote the expansion and shrinking ratios, respectively.
Regions around the boundaries of each proposal are considered uncertain, as the pseudo labels in these areas are expected to contain more noise compared to other regions.

Then we compute the union of all masks as follows:
\begin{equation}
\boldsymbol{U}(t) = \bigcup_{i=1}^{\hat{K}} \mathcal{M}_i(t),
\end{equation}
where $\boldsymbol{U} \in \mathbb{R}^{\hat{T}}$ represents the global uncertainty mask.
Note that the uncertainty mask is generated at multiple scales after assigning proposals to different FPN layers, with $\hat{T}$ representing the total number of anchors across all layers.

During warm-up epochs, the uncertainty mask is used to filter out a portion of the boundary for each proposal, preventing it from involving in the training process.
This uncertainty mask is designed to adapt along with the training process: as the model's confidence in the proposal boundaries increases, the uncertain regions gradually decrease linearly.
As training progresses and the accuracy of the proposal boundaries improves, the mask fades, allowing the entire proposal to be utilized for training, ultimately reducing the influence of noisy labels while enhancing overall model performance.

\subsection{Full Branch}

\noindent\textbf{Training.} We introduce the regression-based model following AFormer~\cite{actionformer} and TriDet~\cite{tridet}.
In the training stage, the regression-based model is designed to leverage multiple priors, including SPs $\boldsymbol{Z}$ and pseudo proposals $\hat{P}$ to enhance performance. 
Sharing the extracted snippets with the weak branch, we first use a transformer encoder to encode the features.
Then a Feature-Pyramid-Network (FPN) is applied to further process the encoded features at multiple scales as $\hat{T}$ anchors.
Then we assign all pseudo proposals from $\hat{P}$ to different scales based on their duration. 
A Classifier Head is applied to predict the class at each anchor as follows:
\begin{align}
    L_{cls} &= \frac{1}{N_{pos}} \sum_{t=1}^{\hat{T}} \mathbbm{1}_{c_t^l > 0 \land \boldsymbol{U}_t = 1} \sigma_{IoU} L_{foc}(1) \nonumber \\
    &\quad + \frac{1}{N_{neg}} \sum_{t=1}^{\hat{T}} \mathbbm{1}_{c_t^l = 0 \land \boldsymbol{U}_t = 1} L_{foc}(0),
\label{equ:cls}
\end{align}
where $N_{pos}$ and $N_{neg}$ denote the number of foreground and background anchors, respectively.
The model is trained only with anchors in certain regions specified by $\boldsymbol{U}_t$. 
We use focal loss $L_{foc}$~\cite{focal} for snippt classification, and $\sigma_{IoU}$ denotes the temporal IoU between the predicted proposal and the pseudo proposal.

For each anchor, the start and end of the proposal are decoded using the Regression Head. The regression loss $L_{reg}$ is calculated as follows:
\begin{equation}
    L_{reg}=\frac{1}{N_{pos}}\sum_{t=1}^{\hat{T}}\mathbbm{1}_{c_t^l>0 \land \boldsymbol{U}_t=1}(1-\sigma_{IoU}),
\label{equ:cls}
\end{equation}
where regression loss is applied only to positive anchors.

Also, to leverage the snippet-level priors and enhance the feature representation, we apply an Attention Head and SoftMax operation for snippet-level classification before FPN. 
We apply $L_{att}$ as follows:
\begin{align}
    L_{att} &= \frac{1}{M} \sum_{t=1}^{T} \sum_{l=1}^{C+1} \mathbbm{1}_{\boldsymbol{Z}_t^l > \tau} \, L_{foc}(l),
\end{align}
where $M$ is the total number of filtered snippets, and $\boldsymbol{Z} \in \mathbb{R}^{C+1}$.
Only snippets with high confidence (above threshold $\tau$) are classified, and those with incorrect labels are filtered with $\boldsymbol{y}$.

By combining all optimization objectives introduced above, we obtain the overall loss function:
\begin{equation}
  L=L_{reg}+L_{cls}+\lambda L_{att},
  \label{equ:finalloss}
\end{equation}
where $\lambda$ is the balancing hyper-parameter of $L_{att}$. 
We train the full branch in a teacher-student manner, following \cite{actionformer,tridet}.
The overall loss $L$ is applied only to the student model for parameter updates, while the teacher model is updated with an exponential moving average (EMA) after warm-up epochs.

\noindent\textbf{Refinement.} During the full branch training stage, refinement is applied to the pseudo proposals and uncertainty mask after the warm-up epoch. Initially, the pseudo proposals $\hat{P}$ are generated by the base model. After the warm-up, the proposals produced by the regression-based model are considered to have more reliable boundaries and confidence scores. 

\begin{table*}[!t]
    \centering
    \resizebox{1.0\linewidth}{!}{
    \begin{tabular}{c|c| c c c c c c c | c c c}
        \toprule 
        \multirow{2}{*}{Supervision} & \multirow{2}{*}{Methods} & \multicolumn{7}{c|}{mAP@IoU($\%$)} & \multicolumn{3}{c}{AVG} \\
        \cmidrule(lr){3-9} \cmidrule(lr){10-12}
         & & 0.1 & 0.2 & 0.3 & 0.4 & 0.5 & 0.6 & 0.7 & (0.1:0.5) & (0.3:0.7) & (0.1:0.7) \\
        \midrule
        \multirow{5}{*}{\shortstack{Fully \\ (TAL)}}
        & A2Net \tiny{TIP'20} & 61.6 & 60.2 & 58.6 & 54.1 & 45.5 & 32.5 & 17.2 & 55.9 & 41.6 & 47.1 \\
        & Tad-MR \tiny{ECCV'20} & - & - & 53.9 & 50.7 & 45.4 & 38.0 & 28.5 & - & 43.3 & - \\
        & TadTR \tiny{TIP'22} & - & - & 74.8 & 69.1 & 60.1 & 46.6 & 32.8 & - & 56.7 & - \\
        & AFormer \tiny{ECCV'22} & - & - & 82.1 & 77.8 & 71.0 & 59.4 & 43.9 & - & 66.8 & - \\
        & TriDet \tiny{CVPR'23} & 87.1 & 86.2 & 83.6 & 80.3 & 73.0 & 62.2 & 46.8 & 82.0 & 69.2 & 74.2 \\
        % & DyFADet \tiny{ECCV'24}& - & - & 84.0 & 80.1 & 72.7&  61.1& 47.9 & - & 69.2 & - \\
        \midrule
        % \multirow{4}{*}{\shortstack{Point\\ (P-TAL)}} & P-GCN \tiny{ICCV’19} & 69.5 & 67.8 & 63.6 & 57.8 & 49.1 & - & - & 61.6 & - & - \\ 
        % & HRPro \tiny{AAAI'24} & \textbf{85.6} & \textbf{81.6} & \textbf{74.3} & \textbf{64.3} & \textbf{52.2} & \textbf{39.8} & \textbf{24.8} & \textbf{71.6} & \textbf{51.1} & \textbf{60.3} \\
        % & TSPNet \tiny{CVPR'24} & 82.3 & 77.6 & 70.1 & 60.0 & 49.4 & 37.6 & 24.5 & 67.9 & 48.3 & 57.4 \\
        % \cmidrule(lr){2-12}
        % & Ours \small{w/ point} & 83.3&78.1&70.9&59.2&48.6&36.8&23.1 & 68.0 & 47.7 & 57.1 \\
        % \midrule
        \multirow{8}{*}{\shortstack{Weakly \\ (WTAL)}}
        & CO2-Net \tiny{MM'21} & 70.1 & 63.6 & 54.5 & 45.7 & 38.3 & 26.4 & 13.4 & 54.4 & 35.7 & 44.6  \\
        & DELU \tiny{ECCV’22} & 71.5 & 66.2 & 56.5 & 47.7 & 40.5 & 27.2 & 15.3 & 56.5 & 37.4 & 46.4 \\
        & PMIL \tiny{CVPR'23} & 71.8 & 67.5 & 58.9 & 49.0 & 40.0 & 27.1 & 15.1 & 57.4 & 38.0 & 47.0 \\
        & Li et al. \tiny{CVPR'23} & - & - & 56.2 & 47.8 & 39.3 & 27.5 & 15.2 & - & 37.2 & -\\
        & PivoTAL \tiny{CVPR'23} & 74.1 & 69.6 & 61.7 & 52.1 & 42.8 & 30.6 & 16.7 & 60.1 & 40.8 & 49.6 \\
        & DDG-Net \tiny{ICCV'23} & 72.5 & 67.7 & 58.2 & 49.0 & 41.4 & 27.6 & 14.8 & 57.9 & 37.9 & 47.3 \\
        & ISSF \tiny{AAAI'24} & 72.4 & 66.9 & 58.4 & 49.7 & 41.8 & 25.5 & 12.8 & 57.8 & 37.6 & 46.8 \\
        \cmidrule(lr){2-12}
        & Ours & \textbf{76.8} & \textbf{72.6} & \textbf{65.4} & \textbf{55.6} & \textbf{44.1} & \textbf{31.3} & \textbf{17.1} & \textbf{62.9} & \textbf{42.7} & \textbf{51.9} \\ 
        \bottomrule 
    \end{tabular}
    }
    \caption{Comparison results with existing methods on THUMOS14 dataset. Our method achieve state-of-the-art WTAL performance with significant improvements.
    % Although other stronger base models could achieve even higher performance, here we use DELU as a stable yet simple base model.
    }
    \label{resultsthumos} 
\end{table*}

In each training iteration, the proposal outputs from the regression model are combined with the original pseudo proposals $\hat{P}$ and updated through RickerFusion. 
This updated set of proposals is then used to refine the uncertainty mask, which guides the model toward more accurate temporal boundaries. 
The parameters $\alpha$ and $\beta$ in the uncertainty mask are linearly degraded as training progresses beyond the warm-up stage. 
This linear degradation reflects our assumption that the proposal boundaries become increasingly reliable and accurate with further training.
Consequently, the model gradually relies less on the uncertainty mask and more on the refined proposals for final predictions.

\noindent \textbf{Inference.} 
% \subsection{Inference}
After the two-stage training, we use the regression-based full branch for inference.
To obtain the video-level classification results, we first apply attention predicted by the attention head to filter the top-k scores for each class, and then average them.
The predicted attention score is then used to filter the video classes, retaining only those proposals within the predicted classes.
Proposals with classes below a certain threshold are discarded.
Finally, we apply soft-NMS~\cite{softnms} to these proposals to obtain the final predictions.
\footnotetext[1]{The reported results use the same hyperparameters as those from ActivityNet1.2 in the original papers.}

\section{Experiment}

\subsection{Datasets and Metrics}
\textbf{Datasets.} We use two commonly used datasets for our experiments: THUMOS14~\cite{thumos}, and ActivityNet1.3~\cite{activitynet}. 
The THUMOS14 dataset consists of 200 validation videos and 213 testing videos, spanning 20 categories, with each video containing an average of 15.4 action instances. 
ActivityNet1.3 contains 10,024 training and 4,926 validation videos across 200 categories.
On average, each video contains 1.6 action instances and about $36\%$ frames are ambiguous action contexts or non-action backgrounds.

\noindent \textbf{Metrics.}  We evaluate the TAL performance with mean Average Precision (mAP) at different Intersections over Union (IoU) thresholds. The IoU thresholds for THUMOS14 are set at [0.1:0.1:0.7], and for ActivityNet1.3 at [0.5:0.05:0.95]. Also, we present results for THUMOS14 at [0.1:0.1:0.5] and [0.3:0.1:0.7].

\noindent \textbf{Implementation Details.} we use DELU~\cite{delu}, DDG-Net~\cite{ddgnet} and CO2-Net~\cite{co2net} as the base model. 
And we follow AFormer~\cite{actionformer} and TriDet~\cite{tridet} to build the regression model.
To extract both RGB and optical flow features, we use Kinetics~\cite{kinetics} pre-trained I3D~\cite{i3d}.
The extracted features have a dimension of 1024 for both streams. 
For the THUMOS14 datasets, following AFormer~\cite{actionformer}, we extract one snippet feature for every 16 frames with a stride 4 (stride 16 for the base model), and for ActivityNet1.3, we extract one snippet feature for every 16 frames, with a video frame rate of 25 fps. 
We set the total number of epochs to 38 for THUMOS14 and 16 for ActivityNet1.3, with warm-up epochs of 20 and 10, respectively.
% $\lambda$ is set to 0.2. 
For FPN, we use 6 layers, and we set $\tau$ to 0.8 and $\lambda$ to 0.2 for $L_{att}$.
The learning rates for THUMOS14 and ActivityNet1.3 are 1e-4 and 1e-3, respectively. 
For the uncertainty mask, we set $\alpha=0.1$ and $\beta=0$ for THUMOS14, and $\alpha=0.05$ and $\beta=0.05$ for ActivityNet1.3.
We conduct our experiments on a single NVIDIA A40 GPU.
%-------------------------------------------------------------------------

\begin{table}[t]
    \centering
    \resizebox{1.0\linewidth}{!}{
    \begin{tabular}{c|c| c c c | c }
        \toprule 
        \multirow{2}{*}{Supervision} & \multirow{2}{*}{Methods} & \multicolumn{3}{c|}{mAP@IoU($\%$)} & \multicolumn{1}{c}{AVG} \\
        \cmidrule(lr){3-5} \cmidrule(lr){6-6}
         & & 0.5 & 0.75 & 0.95 & (0.5:0.95) \\
        \midrule
        \multirow{4}{*}{\shortstack{Fully \\ (TAL)}}
        & A2Net \tiny{TIP'20} & 43.6 & 28.7 & 3.7 & 27.8 \\
        & ReAct \tiny{ECCV'22} & 49.6 & 33.0 & 8.6 & 32.6 \\
        & AFormer \tiny{ECCV'22} & 54.7 & 37.8 & 8.4 & 36.6 \\
        & TriDet \tiny{CVPR'23} & 54.7 & 38.0 & 8.4 & 36.8 \\
        \midrule
        % \multirow{3}{*}{\shortstack{Point\\ (P-TAL)}} 
        % & Yin et al \tiny{BMVC'23} & 48.3 & 27.8 & 7.0 & 29.1 \\
        % & HRPro \tiny{AAAI'24} & 42.8 & 27.2 & \textbf{8.0} & 27.1 \\ 
        % \cmidrule(lr){2-6}
        % & Ours \small{w/ point} & \textbf{50.1} & \textbf{30.4} & 7.8 & \textbf{30.1} \\
        % \midrule
        \multirow{8}{*}{\shortstack{Weakly \\ (WTAL)}}
        & CO2-Net\footnotemark[1] \tiny{MM'21} & 40.8 & 25.6 & 5.7 & 25.6 \\
        & DELU\footnotemark[1] \tiny{ECCV'22} & 40.8 & 26.1 & 6.0 & 25.8\\
        & PMIL \tiny{CVPR'23} & 41.8 & 25.4 & 5.2 & 25.5\\
        & Li et al. \tiny{CVPR'23} & 41.8 & 26.0 & 6.0 & 26.0\\
        & PivoTAL \tiny{CVPR'23} & 45.1 & 28.2 & 5.0 & 28.1 \\
        & DDG-Net\footnotemark[1] \tiny{ICCV'23} & 40.6 & 26.0 & 6.1 & 25.7\\
        & ISSF \tiny{AAAI'24} & 39.4 & 25.8 &  6.4 & 25.8 \\
        \cmidrule(lr){2-6}
        & Ours & \textbf{46.9} & \textbf{28.6} & \textbf{6.5} & \textbf{29.0} \\
        \bottomrule 
    \end{tabular}
    }
    \caption{Comparison results with existing methods on ActivityNet1.3 dataset. Our method also achieve state-of-the-art WTAL performance.}
    \label{resultsanet} 
\end{table}

\subsection{Comparison with State-of-the-art Methods}

In this section, we choose DELU~\cite{delu} as our base model, which is a simple yet effective baseline. Experiments with different baselines will be discussed in our ablation studies.
% Note that there are other more advanced methods~\cite{ddgnet,plateau,atce} as shown in~\cref{generalization}, but we select DELU as a simple yet stable base model.
We compare the performance of PseudoFormer with existing state-of-the-art methods on the THUMOS14 dataset in~\cref{resultsthumos}. 
We also report results from previous representative fully supervised TAL methods for reference.

We observe that PseudoFormer outperforms the base model DELU by $5.5\%$ for average mAP at (0.1:0.7). 
Similar improvements are observed for other mAP scores. 
Compared with the previous SOTA method PivoTAL~\cite{pivotal}, which is also a 2-stage regression-based method, PseudoFormer outperforms it at each IoU threshold ranging from 0.1 to 0.7 by a large margin.
The results of PseudoFormer even outperform early fully supervised TAL methods, marking a significant step towards bridging the gap between WTAL to TAL.
Noting that PseudoFormer can even outperform previous fully supervised method, bridging the gap between WTAL and TAL.

We also compare PseudoFormer with existing methods on the AcitivityNet-v1.3 dataset in~\cref{resultsanet} and observe a similar trend where PseudoFormer outperforms the existing best method PivoTAL by $0.9\%$.
In summary, our PseudoFormer achieves SOTA WTAL performance on both THUMOS14 and AcitivityNet-v1.3 benchmarks.
% And the performance of the proposed method is even surpassing some recent point-supervised methods.

% For PTAL, PeudoFormer simply applies point annotations as a reference to filter noisy proposals in RickerFusion when generating pseudo labels. 
% The proposed method achieves performance close to existing state-of-the-art method HRPro~\cite{hrpro} on THUMOS14, and surpasses the existing SOTA method on ActivityNet1.3.
% Since the average number of proposals per video in ActivityNet1.3 is much smaller than in THUMOS14, point annotations do not yield the same significant improvement as observed in THUMOS14.
% PseudoFormer underperforms recent PTAL methods at higher IoU thresholds due to a lack of specialized design for fully leveraging point annotations to generate accurate pseudo labels.
% However, the results still demonstrate the extensiveness of PseudoFormer.

\subsection{Ablation Study}
\label{ablationstudy}

\noindent\textbf{Effectiveness of the Components.} 
In this section, We perform ablation studies of different components in~\cref{ablation}.
As shown in Exp~1, the base model proposal achieves an average mAP of $46.4\%$.
When directly applying all output proposals (for a snippet within several overlapping proposals, assigning the one with the highest score, details in~\cref{generalization}), there is a $1.8\%$ improvement.
Using RickerFusion to improve the quality of the pseudo labels results in another significant $2.6\%$ improvement in Exp~2.
The two experiments demonstrate the effectiveness of the full branch framework and the significance of the quality of pseudo labels for performance.
Introducing the attention loss in Exp~3 $L_{att}$ enables PseudoFormer to leverage snippet-level priors, contributing an additional $0.5\%$ improvement. 
This loss function enhances video-level classification and aids in optimizing encoder features before the feature pyramid network (FPN).
For handling noisy labels, in Exp~4$\&$5 the uncertainty mask yields a $0.3\%$ gain, while label refinement provides a further $0.3\%$ boost. 
These two modules help the model gradually update with better pseudo labels throughout the training process.
With all modules integrated in Exp~6, PseudoFormer achieves a final performance of $51.9\%$, setting a new state-of-the-art.

\begin{table}[t]
    \centering 
    \resizebox{1.0\linewidth}{!}{
    \begin{tabular}{c | c c c c c | c }
        \toprule 
        Exp & \makecell{Regression\\Model} & \makecell{Ricker\\Fusion} & \makecell{Attention\\Loss} & \makecell{Uncertainty\\Mask} & \makecell{Label\\Refinement} & \makecell{AVG\\ (0.1:0.7)} \\
        \midrule
        1 & \ding{55} & \ding{55} &  \ding{55} & \ding{55} & \ding{55} & 46.4 \\
        2 & \checkmark & \ding{55} &  \ding{55} & \ding{55} & \ding{55} & 48.2 \\
        3 & \checkmark & \checkmark &  \ding{55} & \ding{55} & \ding{55} & 50.8 \\
        4 & \checkmark & \checkmark & \checkmark & \ding{55} & \ding{55} & 51.3 \\
        5 & \checkmark & \checkmark & \checkmark & \checkmark & \ding{55} & 51.6 \\
        \midrule
        6 & \checkmark & \checkmark & \checkmark & \checkmark & \checkmark & 51.9 \\
        \bottomrule 
    \end{tabular}
    }
    \caption{Ablation study on each component of PseudoFormer.}
    \label{ablation}
\end{table}
\begin{table}[t]
    \centering
    \resizebox{1.0\linewidth}{!}{
    \begin{tabular}{c| c c c c c c c | c }
        \toprule 
        \multirow{2}{*}{Methods} & \multicolumn{7}{c|}{mAP@IoU($\%$)} & AVG \\
        \cmidrule(lr){2-8} \cmidrule(lr){9-9}
        & 0.1 & 0.2 & 0.3 & 0.4 & 0.5 & 0.6 & 0.7 & (0.1:0.7) \\
        \midrule
        Hard  &  72.1 & 66.7 &  58.2 & 48.0  & 36.4 & 25.9  & 14.7  & 46.0 \\
        Soft  & 75.9  &  70.2 & 61.8  & 50.9  & 38.7  & 25.6  & 14.6 & 48.2 \\
        Top-K & 75.5 & 69.7 & 60.9 & 50.0 & 38.1 & 26.4 & 15.2 & 47.3 \\
        Threshold & 75.0  & 69.9  & 61.4  &  51.6 &  40.6 & 28.1  & 16.3  & 49.0 \\
        Gauss & 75.4 & 70.1 & 61.6 & 51.0 & 40.1 & 27.6 & 15.8 & 49.2  \\
        \midrule
        RickerFusion  & \textbf{76.3} & \textbf{71.8} & \textbf{63.6} & \textbf{53.8} & \textbf{42.5} & \textbf{29.7} & \textbf{16.5} & \textbf{50.8} \\
        \bottomrule
    \end{tabular}
    }
    \caption{Comparison results between different pseudo label generation strategies with RickerFusion.}
    \label{generationstrategy}
\end{table}

\noindent\textbf{Different Fusion Methods.}
As shown in~\cref{ablationstudy}, we observe that the primary improvement comes with RickerFusion.
This demonstrates that the quality of the pseudo proposals is crucial for the final performance.
We compare RickerFusion with different fusion methods. 
As shown in~\cref{generationstrategy}, we conduct different fusion methods to improve the quality of output proposals from the base model after soft-NMS.
(1) \emph{Hard}: For each snippet, assign it to only one proposal or background, filtering redundant overlapping proposals based on confidence scores.
(2) \emph{Soft}: Keep all proposals as input. If a snippet belongs to multiple overlapping proposals, assign it to the one with the highest confidence score. This strategy serves as the baseline in~\cref{ablationstudy}.
(3) \emph{Top-K}: For each video, select the top-K proposals with the highest confidence scores.
(4) \emph{Threshold}: Filter proposals directly based on a confidence threshold.
(5) \emph{Gauss}: Following~\cite{gaussfusion}, assume proposal boundaries and confidence scores follow a Gaussian distribution. We group proposals by IoU with the top proposal and compute the expected boundaries using the scores.

From~\cref{generationstrategy}, We observe that RickerFusion achieves the best performance. 
Unlike the Hard, Soft, Top-K, and Threshold strategies, RickerFusion has a light reliance on hyper-parameters.
For the Gaussian strategy, pre-setting the number of proposal groups based on IoU can lead to either over-segmentation or the omission of important proposals, which reduces the overall accuracy of the full branch model.
The RickerFusion emphases the centers of the proposals, while suppresses the regions out of the boundaries, achieving the best results.

\noindent\textbf{Different Base Models.}
\label{generalization}
The proposed method focuses on utilizing generated information from the base model and effective training strategies, which are agnostic to the structure of the base locator.
Consequently, PseudoFormer can be easily integrated into any Multiple Instance Learning (MIL) based WTAL method.
To evaluate the generalization capability of PseudoFormer across different baselines, we employ our approach on three WTAL methods: CO2-Net~\cite{co2net}, DELU~\cite{delu}, and DDG-Net~\cite{ddgnet}. 
The results in~\cref{basemodel} demonstrate significant performance improvements of $6.6\%$, $5.5\%$, and $5.1\%$ for the 3 base models when integrated with PseudoFormer.
With a weaker baseline (44.6 vs 46.4), our method can still achieve similar final performance (51.2 vs 51.9). This indicates that the pseudo label generation and training pipeline is the key of our performance gain.
Besides, we observe that our approach exhibits remarkable results for a strong base model DDG-Net and surpasses the current SOTA PivoTAL~\cite{pivotal} by $2.8\%$.
% Also, we note that with a stronger base model, there are fewer improvements.
% Our results suggest that for stronger models, in which the outputs are already highly accurate, there is less room for further improvement.
This demonstrates the generalizability of our approach.

% 76.6 71.6 64.6 55.1 43.4 31.1 17.0 51.3

\begin{table}[t]
    \centering
    \resizebox{1.0\linewidth}{!}{
    \begin{tabular}{c| c c c c c c c | c }
        \toprule 
        \multirow{2}{*}{Methods} & \multicolumn{7}{c|}{mAP@IoU($\%$)} & AVG \\
        \cmidrule(lr){2-8} \cmidrule(lr){9-9}
        & 0.1 & 0.2 & 0.3 & 0.4 & 0.5 & 0.6 & 0.7 & (0.1:0.7) \\
        \midrule
        CO2-Net & 70.1 & 63.6 & 54.5 & 45.7 & 38.3 & 26.4 & 13.4 & 44.6\\
        + PseudoFormer & \textbf{76.6} & \textbf{71.6} & \textbf{64.6} & \textbf{55.1} & \textbf{43.5} & \textbf{31.1} & \textbf{16.9} & $\mathbf{51.2_{\uparrow 6.6}}$\\
        \midrule
        DELU & 71.5 & 66.2 & 56.5 & 47.7 & 40.5 & 27.2 & 15.3 & 46.4\\
        + PseudoFormer & \textbf{76.8} & \textbf{72.6} & \textbf{65.4} & \textbf{55.6} & \textbf{44.1} & \textbf{31.3} & \textbf{17.1} & $\mathbf{51.9_{\uparrow 5.5}}$\\
        \midrule
        DDG-Net & 72.5 & 67.7 & 58.2 & 49.0 & 41.4 & 27.6 & 14.8 & 47.3 \\
        + PseudoFormer & \textbf{77.2} & \textbf{72.2} & \textbf{65.8} & \textbf{56.8} & \textbf{44.8} & \textbf{31.0} & \textbf{18.4} & $\mathbf{52.4_{\uparrow 5.1}}$ \\
        \bottomrule
    \end{tabular}
    }
    \caption{Results of PseudoFormer into different multiple instance learning base methods on THUMOS14 dataset.}
    \label{basemodel}
\end{table}

\subsection{Qualitative Analysis}
\begin{figure}
  \centering
       \centering
\includegraphics[width=0.95\linewidth]{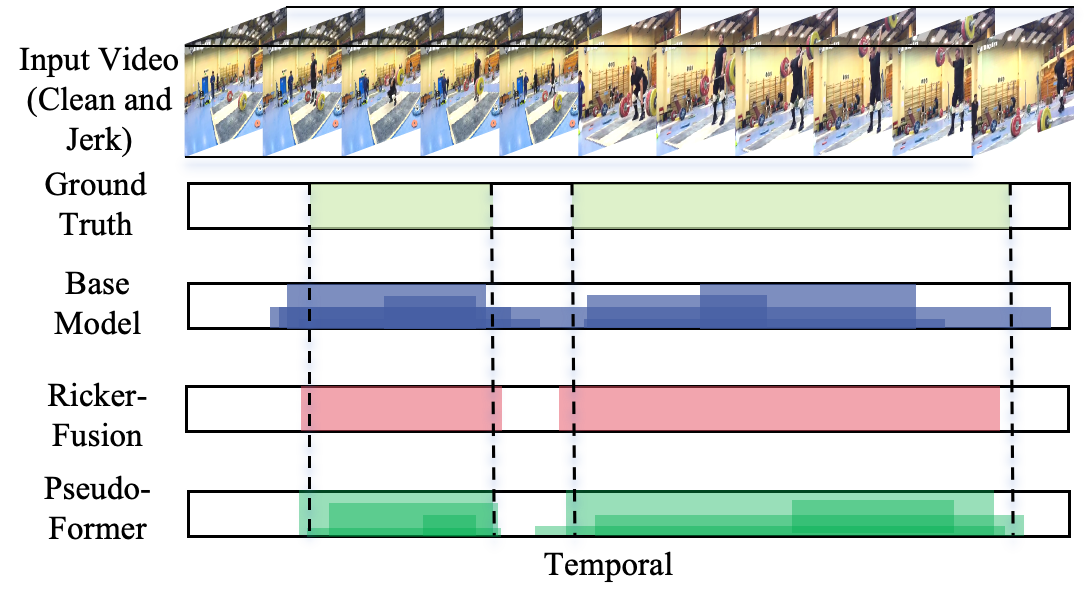}
       \caption{Visualization for ground truth on a test video, base model (DELU), after RickerFusion and PseudoFormer. For Base Model and PseudoFormer, we visualize the top-4 proposals overlapping the ground truth. With the input video, PseudoFormer produces more consistent and accurate predictions.}
       \label{fig:visualization}
\end{figure}

We visualize some predicted action proposals in a THUMOS14 test video along the temporal axis in~\cref{fig:visualization}, and compare the result with the base model DELU~\cite{delu}.
Additionally, We visualize the pseudo labels to train the regression model after RickerFusion.
For the base model, the output proposals are relatively inaccurate and scattered.
The base model's multi-scale thresholding in post-processing the classification attention sequence (CAS) results in imprecise boundaries.
In contrast, PseudoFormer, as a regression-based method, produces the output proposals with more accurate and consistent boundaries.
The regression-based model does not rely on artificial post-processing.
As shown in~\cref{fig:visualization}, RickerFusion enhances proposal quality from the base model, eliminates noise, and ensures each snippet belongs to one proposal or background, improving the quality of pseudo proposals.

\section{Conclusion}

In this paper, we propose PseudoFormer, a two-branch framework that achieves strong results on both WTAL and PTAL tasks. 
We address the three primary challenges in bridging the gap in both framework and performance between Weakly-supervised and Fully-supervised Temporal Action Localization (TAL).
Our method introduces a powerful pseudo-label generation strategy, leveraging RickerFusion to generate pseudo proposals with precise boundaries. 
By training both snippet-level and proposal-level labels, PseudoFormer enhances learning, capturing different priors from the base model. 
Additionally, the uncertainty mask and label refinement mechanisms effectively handle noisy labels, ensuring robust model training.
This work establishes a foundation for future research in Temporal Action Localization, paving the way toward a unified framework adaptable to varying levels of supervision.

\section*{Acknowledgment}
This work was supported by Beijing Natural Science Foundation under Grant 4244082, and National Natural Science Foundation of China under Grant 62402034.  
The authors would like to thank Prof. Junsong Yuan, and Dr. Yuanhao Zhai for useful discussions.

\clearpage

{
    \small
    \bibliographystyle{ieeenat_fullname}
    \bibliography{refer}
}
\end{document}